\documentclass{article}
\usepackage[nonatbib,preprint]{neurips}
\usepackage[numbers,sort&compress]{natbib}
\usepackage{subfig}
\usepackage{listings}
\usepackage{marvosym}

\usepackage[export]{adjustbox}
\usepackage[ruled]{algorithm2e}
\usepackage[inline, shortlabels]{enumitem}
\usepackage[T1]{fontenc}
\usepackage{hyperref}
\usepackage{microtype}
\usepackage{pifont}
\usepackage{xurl}
\usepackage{graphicx}
\usepackage{booktabs}
\usepackage{tabularray}
\usepackage{listings}
\usepackage{amsmath, amsfonts}
\usepackage{nicefrac}
\usepackage{wrapfig}
\usepackage{amsmath}
\usepackage{multirow}
\usepackage{pifont}
\usepackage{wasysym}
\usepackage{amssymb}
\UseTblrLibrary{booktabs}

\makeatletter
\newcommand{\ssymbol}[1]{\@fnsymbol{#1}}
\newcommand{\romanNumeral}[1]{\expandafter\@slowromancap\romannumeral #1@}
\makeatother

\usepackage{physics}
\usepackage{mathtools}
\usepackage[table]{xcolor}
\usepackage{mdframed}
\hypersetup{
    colorlinks=true,
    citecolor={rgb,255:red,26;green,134;blue,163},
    colorlinks=true,
    linkcolor={rgb,255:red,255;green,99;blue,71} 
}

\definecolor{teagreen}{HTML}{e1f8bd}
\definecolor{lightblue}{HTML}{d3f4ff}
\definecolor{front-color}{HTML}{ffe9b8}
\definecolor{lightpink}{HTML}{ebdef0}
\definecolor{Gray}{gray}{0.7}

\definecolor{Magenta}{rgb}{0.8, 0.1, 0.6}            
\title{AV-Reasoner: Improving and Benchmarking Clue-Grounded Audio-Visual Counting for MLLMs}

\author{
\textbf{Lidong Lu}\textsuperscript{*},
\textbf{Guo Chen}\textsuperscript{*},
\textbf{Zhiqi Li}\textsuperscript{},
\textbf{Yicheng Liu}\textsuperscript{},
\textbf{Tong Lu}\textsuperscript{\Letter}
\vspace{0.5em}
\\
Nanjing University\\
\href{https://av-reasoner.github.io} {\path{https://av-reasoner.github.io}}
}

\begin{document}
\makeatletter
\renewcommand{\@makefntext}[1]{%
  \parindent 0pt  
  \noindent\makebox[0em][r]{\@thefnmark}~#1}
\makeatother

\begingroup
\renewcommand{\thefootnote}{}
\footnotetext{\textsuperscript{*}Equal contribution. \textsuperscript{\Letter}Corresponding author.}
\endgroup
\maketitle

\begin{abstract}
  Despite progress in video understanding, current MLLMs struggle with counting tasks. Existing benchmarks are limited by short videos, close-set queries, lack of clue annotations, and weak multimodal coverage. In this paper, we introduce CG-AV-Counting, a manually-annotated clue-grounded counting benchmark with 1,027 multimodal questions and 5,845 annotated clues over 497 long videos. It supports both black-box and white-box evaluation, serving as a comprehensive testbed for both end-to-end and reasoning-based counting. To explore ways to improve model's counting capability, we propose AV-Reasoner, a model trained with GRPO and curriculum learning to generalize counting ability from related tasks. AV-Reasoner achieves state-of-the-art results across multiple benchmarks, demonstrating the effectiveness of reinforcement learning. However, experiments show that on out-of-domain benchmarks, reasoning in the language space fails to bring performance gains.
\end{abstract}

\section{Introduction}

Counting is a crucial indicator of fine-grained alignment and reasoning in Multimodal Large Language Models (MLLMs). Unlike coarse-grained tasks, counting requires precise temporal and spatial grounding. Models must detect, localize, and accumulate instances across frames or scenes, often under challenging visual conditions. This makes it a valuable proxy task for evaluating a model's ability to align vision, audio and language at a detailed level. Moreover, counting has broad practical applications across a variety of real-world scenarios, including crowd analysis, object inventory, and repetitive action recognition. 

Despite recent progress in tasks like video question answering, temporal grounding, etc.~\cite{bai2025qwen2,zhu2025internvl3,chen2025eagle,zhang2025videollama,videollama,cheng2024videollama,videollava,chen2023internvl,chen2024internvl2,wang2022internvideo,wang2024internvideo2,wang2024cogvlm,hong2024cogvlm2,chen2023videollm}, MLLMs' performance on counting task remains limited. 
Besides, existing counting benchmarks often suffer from the following shortcomings:
\begin{itemize}[leftmargin=4.5mm, itemsep=0pt, topsep=0pt]
\item \emph{Short Video Duration}: Most benchmarks~\cite{dwibedi2024ovr,video-r1_2025,dwibedi2020counting,zhang2021repetitive,hu2022transrac} focus on videos shorter than one minute, which restricts the evaluation of counting and grounding performance over longer temporal contexts.

\item \emph{Closed-Set Queries}: Most benchmarks have a close set of queries~\cite{dwibedi2020counting,zhang2021repetitive,hu2022transrac}, limiting their generalization. Although some recent benchmarks support open-vocabulary queries~\cite{dwibedi2024ovr}, their use of short video clips makes it easy for models to exploit shortcuts.

\item \emph{Lack of Clue Annotations}: Existing datasets typically lack annotations for counting clues~\cite{video-r1_2025,dwibedi2020counting,zhang2021repetitive}, making it difficult to assess whether models perform genuine reasoning or rely on guesswork.

\item \emph{Single-Modality Evaluation}: With the emergence of omni-modal MLLMs, most benchmarks still focus solely on visual inputs, failing to test how these models process and combine information from different modalities.
\end{itemize}

To address these limitations, we propose CG-AV-Counting, a manually-annotated clue-grounded benchmark for evaluating MLLMs' counting ability in long videos with multimodal queries. Based on 497 videos from CG-Bench~\cite{chen2024cg} spanning over ten content categories, our dataset includes 1,027 annotated questions across five reference-query modalities: visual-only, audio-only, visual-reference audio-query, audio-reference visual-query, and joint audio-visual. Counting targets cover events, objects, and attributes (e.g., gender, color), each with annotated reference intervals and couting clues. Besides, we design both white-box and black-box evaluations to assess both end-to-end and reasoning-based counting.

We evaluate a range of closed-source and open-source MLLMs using this benchmark. The result shows that most models struggle with complex video counting tasks, especially under white-box evaluation, and open-source audio-visual models often perform worse than vision-only models. It indicates that current MLLMs still have significant room for improvement in this task.

In an effort to explore strategies for enhancing counting ability, and given the lack of counting-specific training data, we propose a capability transfer approach. Based on Ola-Omni~\cite{liu2025ola}, we develop AV-Reasoner, trained with GRPO~\cite{guo2025deepseek} and curriculum learning~\cite{narvekar2020curriculum}. Instead of relying solely on limited counting annotations, AV-Reasoner progressively learns tasks closely related to counting, such as audio-visual understanding, temporal grounding, and spatial grounding, enabling it to acquire counting ability through ability generalization. This approach achieves SOTA performance on general audio-visual understanding tasks such as Audio Visual Question Answering (AVQA), Audio Visual Temporal Grounding (AVTG), and Audio Refered Image Grounding (ARIG), and significantly improves counting performance compared to the base model.

\begin{figure}
  \centering
  \includegraphics[width=\linewidth]{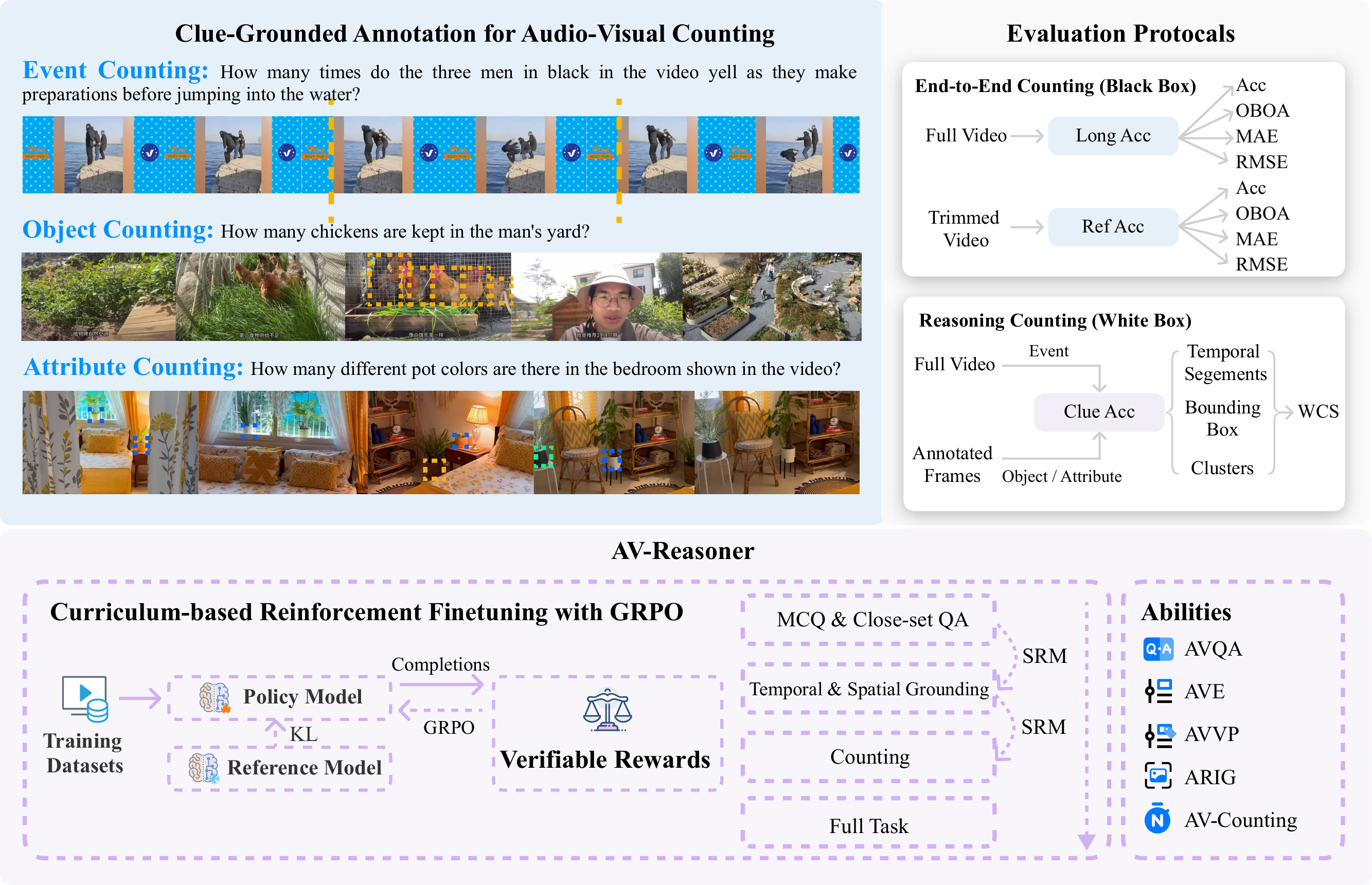}
  \caption{\textbf{Overview of CG-AV-Counting benchmark and AV-Reasoner baseline.} Our benchmark covers three types of counting targets, providing end-to-end counting evaluation based on black-box testing and reasoing counting evaluation based on white-box testing, which can comprehensively evaluate the accuracy and interpretability of the model counts.AV-Reasoner uses GRPO algorithms based on course learning for intensive fine-tuning, introduces the stage review mechanism (SRM) and the Full Task RL to alleviate forgetting between tasks. AV-Reasoner has various capabilities such as AVQA, AVE, AVVP, ARIG, AV-Counting, etc.}
  \label{fig: teaser}
  \vspace{-5mm}
\end{figure}

\section{Related Work}
\textbf{Omni-modal Large Language Models.} Omni-MLLMs are designed to process diverse inputs, such as text, images, audio, and video, within a unified framework. Models like AnyGPT~\cite{zhan2024anygpt} and Unified-IO 2~\cite{lu2024unified} map all modalities to a shared feature space, while Mini-Omni2~\cite{xie2024mini}, OMCAT~\cite{goel2024omcat}, Baichuan-Omni~\cite{li2025baichuan}, Ola-Omni~\cite{liu2025ola}, Crab~\cite{du2025crab}, and VideoLLaMA 2~\cite{cheng2024videollama} use addtional encoders to inject new modalities. For audio-video alignment, Video-SALMONN~\cite{sun2024video} utilizes fine-grained synchronization, Meerkat~\cite{chowdhury2024meerkat} employs optimal transport for feature matching, and PAVE~\cite{liu2025pavepatchingadaptingvideo} integrates audio into visual features using a lightweight module. 
In this paper, we introduce AV-Reasoner, an enhanced version of Ola-Omni~\cite{liu2025ola} that improves counting performance through GRPO training strategy. Beyond counting, AV-Reasoner also achieves new SOTA results on most audio-visual understanding tasks.

\textbf{Video Counting Benchmarks.}  To the best of our knowledge, there is currently no comprehensive benchmark specifically designed to evaluate MLLMs' video counting capabilities. DVD-Counting~\cite{video-r1_2025} and VideoNIAH~\cite{zhao2024needle} use synthetic data for object counting. They have limited counting target variety and do not have long videos. Other benchmarks, such as MVBench~\cite{li2024mvbench} and WorldSense~\cite{hong2025worldsense}, include real-world videos, but counting is only a subtask of the overall evaluation, resulting in a smaller number of samples. Datasets for repetitive action counting~\cite{dwibedi2024ovr,zhang2021repetitive,hu2022transrac,dwibedi2020counting} feature short videos and simple queries, making them unsuitable for evaluating MLLMs. Besides, most benchmarks only have visual queries, which limits their ability to fully evaluate Omni-MLLMs.

To fill these gaps, as is shown in Tab.~\ref{tab:tab1}, we introduce CG-AV-Counting, a benchmark focused on counting tasks in videos longer than 10 minutes. Unlike previous datasets, it has both audio and visual modalities in the queries. The benchmark includes 1,027 samples with various counting targets like objects, events, and attributes. It also provide fine-grained clue annotations for more interpretable evaluation of MLLMs.

\textbf{Large Language Model Reasoning.} Recent advancements in LLM reasoning have focused on improving complex problem-solving abilities. Chain-of-Thought (CoT)~\cite{wei2022chain} helps models break down complex tasks into logical steps, improving performance in arithmetic, commonsense, and symbolic reasoning. Unlike previous supervised methods, DeepSeek R1~\cite{guo2025deepseek} introduces a GRPO-based reinforcement learning approach, guiding models to optimize reasoning strategies without relying on step-by-step annotations. Inspired by this, GRPO-based MLLMs, like Visual-RFT~\cite{liu2025visual}, Video-R1~\cite{feng2025video}, and Video-Chat-R1~\cite{li2025videochat}, have achieved significant results in visual reasoning tasks.

\section{CG-AV-Counting}
\subsection{Dataset Construction}

\begin{figure}
  \centering
  \includegraphics[width=\linewidth]{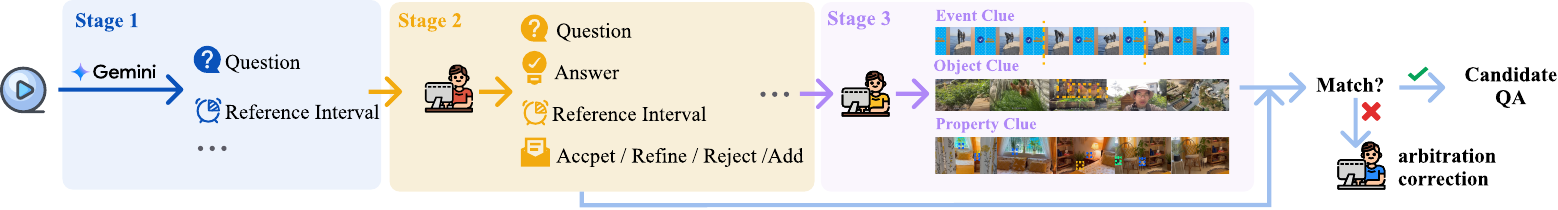}
  \caption{\textbf{A Three-Stage Data Annotation Pipeline for CG-AV-Counting.} It is designed to improve both annotation efficiency and accuracy.}
  \label{fig: dataset}
\end{figure}

As shown in Fig.~\ref{fig: dataset}, the construction of CG-AV-Counting follows three stages:

In the \textbf{question proposal stage}, Gemini 2.0 Flash~\cite{team2023gemini} is used to generate candidate questions and query intervals for each video. Annotators use these as references to improve efficiency.

During the \textbf{answer annotation stage}, human annotators first preview the whole video, select and refine questions that can be grounded globally, draft answers, and determine the interval to localize the query (reference interval). If needed, they can create new questions.

The \textbf{clue annotation stage} provides fine-grained annotations to verify the draft answers and evaluate model's interpretable coutning ability. Human annotators first label counting clues from the video, which is categorized into three types according to the counting target:

\begin{itemize}[leftmargin=4.5mm, itemsep=0pt, topsep=0pt]

\item For event counting, they annotate the start and end timestamps of each event.

\item For object counting, they annotate bounding boxes when objects appear simultaneously or annotate the first appearance of each object. This approach aligns with intuitive human counting practices and reduces annotation errors.

\item For attribute counting, they first perform object counting clue annotation and then group objects with the same query attribute.

\end{itemize}

Finally, the clues are compared with the answer drafted in stage 2. If they match, the corresponding sample is added to the candidate pool, and the final reference interval is determined by combining the draft interval with the clue intervals. Otherwise, arbitration is conducted to decide whether the sample should be retained.

\subsection{Dataset Statistics}
\begin{figure}
  \centering
  \includegraphics[width=\linewidth]{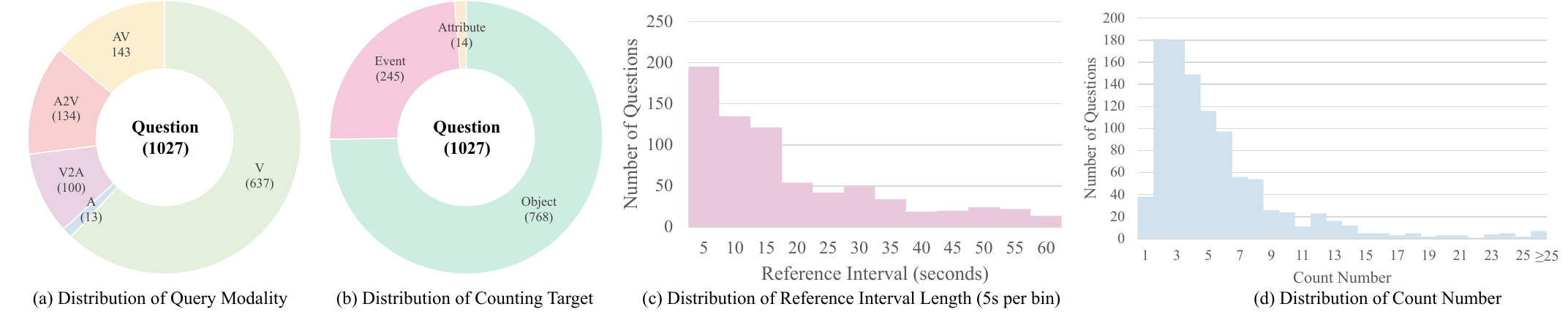}
  \caption{\textbf{Statistics of Our Proposed Benchmark.} Our benchmark features diverse query modalities and a wide range of counting targets. The count distribution follows a long-tail pattern, providing a challenging testbed for evaluating models' generalization and compositional reasoning abilities.}
  \label{fig: dataset2}
\end{figure}
As shown in Fig.~\ref{fig: dataset2}, CG-AV-Counting is based on a subset of 497 videos from CG-Bench~\cite{chen2024cg}. The benchmark includes 1,027 multimodal-query questions and 5,845 fine-grained manually-annotated clue annotations. Nearly 40\% of the samples require the model to use both audio and visual modalities for counting, while others only require the visual modality. This design ensures that the benchmark is applicable to both visual models and audio-visual models. The benchmark includes object, event, and attribute counting target. Among them, attribute counting is more challenging because it requires grouping objects with the same attribute based on the query.

This benchmark spans a numerical range from 1 to 76, with a long-tail distribution, where most counts fall between 1 and 20. Video content includes over 10 categories, such as sports, life record, humor, tutorials, etc., offering greater domain diversity than existing benchmarks. All videos in the benchmark exceed 10 minutes, and reference intervals range from seconds to minutes, covering both short-term and long-range dependencies. 

In summary, as shown in Tab.~\ref{tab:tab1}, unlike previous counting benchmarks, our benchmark incorporates both audio and visual modalities, features more complex queries, and provides fine-grained counting clues to jointly evaluate models' abilities in both end-to-end and reasoning-based counting.

\subsection{Evaluation Metrics}
We follow CG-Bench's settings and use a dual evaluation protocol. It includes \textbf{black-box} and \textbf{white-box} evaluations to comprehensively assess MLLMs' counting ability.

\textbf{Black-box Evaluation} assesses model's end-to-end counting ability under two settings:

\begin{itemize}[leftmargin=4.5mm, itemsep=0pt, topsep=0pt]
\item Long Acc: The model is provided with the entire video and must perform both temporal localization and counting, reflecting its holistic reasoning ability.

\item Ref Acc: The model is given a trimmed segment based on the reference interval, isolating its counting ability from temporal localization.
\end{itemize}

For both settings, we use four complementary metrics to measure counting ability:

\begin{itemize}[leftmargin=4.5mm, itemsep=0pt, topsep=0pt]
\item Accuracy (Acc): It evaluates the model's ability to predict the exact number of instances, reflecting strict counting precision.

\item  Off-By-One Accuracy (OBOA): It allows a small margin of error in counting. A prediction is considered correct if it is off by at most one.

\item  Mean Absolute Error (MAE): It quantifies the average size of counting errors, providing a direct measure of overall counting consistency.

\item Root Mean Square Error (RMSE): It penalizes larger deviations more heavily, highlighting the model's stability and reliability in avoiding severe counting mistakes.
\end{itemize}

\textbf{White-box Evaluation} assesses the model's ability to localize counting targets with explicit evidence: for \textbf{event counting}, the model predicts temporal segments for each event. We use temporal IoU (tIoU) to measure alignment with ground-truth intervals. For \textbf{object counting}, the model is given specific video frames and must locate the first appearance of each object using bounding boxes. Spatial alignment is measured by Intersection over Union (IoU). For \textbf{attribute counting}, predicted bounding boxes are clustered according to the query, and each predicted cluster is evaluated against the corresponding ground-truth cluster using IoU.

To account for both localization and count accuracy, we add a counting accuracy penalty for deviations from the ground-truth. So the \textbf{White-box Counting Score (WCS)} is defined as:

\begin{equation}
\begin{aligned}
\text{WCS} &= \frac{1}{K} \sum_{k=1}^{K} \sqrt{\text{LA}_k \times \text{CAP}_k} \times 100\% \\
\text{LA}_k &= \frac{1}{|\text{GT}_k|} \sum_{j=1}^{|\text{GT}_k|} \text{IoU}(\text{Pred}_k, \text{GT}_k) \\
\text{CAP}_k &= \max\left\{ 0, 1 - \frac{\left| |\text{Pred}_k| - |\text{GT}_k| \right|}{|\text{GT}_k|} \right\}
\end{aligned}
\end{equation}

where $K$ is the number of instance clusters: one for event and object counting, and the number of attribute types for attribute counting. IoU refers to tIoU or IoU depending on the task. 

\(\text{LA}_k\) (Localization Accuracy) measures the average overlap between predicted and ground-truth instances within the \(k\)-th cluster, reflecting how accurately the model localizes instances. To fairly evaluate this metric, a greedy matching algorithm is used to establish an optimal one-to-one correspondence between predicted and ground-truth instances, maximizing the total IoU even when their numbers or order differ.

\(\text{CAP}_k\) (Counting Accuracy Penalty) penalizes the discrepancy between the number of predicted instances and ground-truth instances in the \(k\)-th cluster. It scales from 0 to 1, with a value of 1 indicating perfect count matching and values closer to 0 indicating larger count deviations.

The score increases when predictions are both well-localized and correctly counted, and drops as the count error or localization error grows. It is bounded between 0 and 100. A score of 100 indicates perfect performance, where all predicted instances are accurately localized and the predicted count exactly matches the ground-truth. In contrast, a score of 0 reflects either a severe count mismatch, where the absolute error exceeds the ground-truth count, or a failure to follow the required output format. In practice, when the counting accuracy is low, the penalty tends to dominate the overall score, resulting in a sharp drop regardless of localization quality. Conversely, when the predicted count is close to the ground-truth, the localization accuracy becomes the main factor influencing the final score.

In addition, we report \textbf{Instruction-Following Accuracy (IFA)} to measure the proportion of samples for which the model's output matches the required format. This metric complements WCS by explicitly evaluating the model's ability to follow task-specific output instructions, which is essential for reliable and interpretable predictions.

\subsection{Comparison with Existing Video Counting Benchmarks}

\begin{table*}[t]
\centering
\caption{\textbf{Comparison with existing video counting benchmarks.} "\ding{51}/\ding{55}" indicates partial satisfaction. "\#Test Samples" indicates the number of samples related to the counting task within each benchmark.}
\resizebox{\columnwidth}{!}{%

\begin{tabular}{ccc|cc|cc|ccc}
\toprule
Benchmark   & \#Test Samples & Modaility & Real scene & Long Video & Complex Query & Clue & Event & Object & Attribute \\
\midrule
  DVD-Counting~\cite{video-r1_2025}  & 200 & V & \ding{55} & \ding{55} & \ding{51} & \ding{55} & \ding{55} & \ding{51} & \ding{55}\\ 
  VideoNIAH~\cite{zhao2024needle}  & 450 & V & \ding{55} & \ding{55} & \ding{51} & \ding{55}& \ding{55} & \ding{51} & \ding{55}\\ 
  MVBench~\cite{li2024mvbench}  & 400 & V & \ding{51}/\ding{55} & \ding{51}/\ding{55} & \ding{51} & \ding{55} & \ding{51} & \ding{55} & \ding{55}\\ 
  RepCount~\cite{hu2022transrac}  & 152 & V & \ding{51} & \ding{51} &  \ding{55} &\ding{51} & \ding{51} & \ding{55} & \ding{55}\\ 
  CountixAV~\cite{zhang2021repetitive}  & 563 & A+V & \ding{51} & \ding{55} & \ding{55} & \ding{55}& \ding{51} & \ding{55} & \ding{55}\\
  OVR~\cite{dwibedi2024ovr}  & 12452 & V & \ding{51} & \ding{55} &\ding{51}& \ding{55} & \ding{51} & \ding{55} & \ding{55}\\
  WorldSense~\cite{hong2025worldsense} & 476 & A+V & \ding{51} & \ding{51} &\ding{51}& \ding{55} & \ding{51} & \ding{51} & \ding{55}\\
  CG-AV-Conuting (Ours)  & 1027 & A+V & \ding{51} & \ding{51}&\ding{51} & \ding{51} & \ding{51} & \ding{51} & \ding{51}\\
\bottomrule
\end{tabular}

}
\label{tab:tab1}
\end{table*}

As summarized in Table~\ref{tab:tab1}, existing video counting benchmarks typically suffer from limited modality, content diversity, and reasoning complexity.

Most prior datasets, such as DVD-Counting~\cite{video-r1_2025}, VideoNIAH~\cite{zhao2024needle}, and MVBench~\cite{li2024mvbench}, only contain visual-only samples. In contrast, CG-AV-Counting introduces a richer query structure with audio-visual interactions, supporting audio-referenced visual queries, visual-referenced audio queries, and joint audio-visual counting—enabling evaluation of complex multimodal reasoning scenarios.

Second, regarding video length, most existing benchmarks rely on short clips (typically under 1 minute)~\cite{zhao2024needle,li2024mvbench,video-r1_2025,zhang2021repetitive,dwibedi2024ovr}. CG-AV-Counting leverages long-form videos (all exceeding 10 minutes), requiring sustained temporal reasoning and long-range clue localization.

Third, in terms of counting targets, previous datasets mostly focus on object or event counting~\cite{zhao2024needle,li2024mvbench,video-r1_2025,hu2022transrac,zhang2021repetitive,dwibedi2024ovr}. In contrast, CG-AV-Counting comprehensively covers object, event, and attribute counting, enabling a more fine-grained and versatile evaluation.

Fourth, while prior datasets~\cite{zhao2024needle,li2024mvbench,video-r1_2025,zhang2021repetitive,hong2025worldsense} typically only provide final count labels, CG-AV-Counting includes manually annotated fine-grained counting clues that clearly indicate where and how evidence for the count appears across modalities and time. These annotations not only improve dataset transparency, but also support interpretable and diagnostic evaluation of model behavior.

Finally, beyond standard black-box evaluation of end-to-end counting accuracy, CG-AV-Counting introduces white-box evaluation protocols to assess models’ intermediate reasoning steps. This dual evaluation protocols allow for a more comprehensive and explainable assessment of multimodal counting capabilities.

Overall, CG-AV-Counting significantly broadens the scope and realism of video counting evaluation, establishing a more challenging and representative bnchmark for future multimodal reasoning research.

\subsection{Evaluation Results}
As shown in Tab.~\ref{tab:benchmark}, all models struggle with the counting task, significantly underperforming compared to human performance. Close-source MLLMs generally outperform open-source ones across most metrics. However, even the best-performing close-source models fall far short of human-level accuracy, indicating that accurate counting remains a major challenge for current MLLMs.

Notably, Gemini 2.5 Pro~\cite{team2023gemini} and Gemini 2.5 Flash~\cite{team2023gemini}, which utilize both audio and visual modalities, achieve the strongest results, which reflects their relative robustness across different evaluation settings. Among open-source models, Eagle2.5-8B~\cite{chen2025eagle} is a rare outlier, outperforming several close-source models.

Nevertheless, despite the inclusion of audio-visual samples in the benchmark, the improvement from A+V models is inconsistent. For example, open-source audio-visual models such as UnifiedIO-2 XXL~\cite{lu2024unified} and VideoLLaMA2.1-7B-AV~\cite{cheng2024videollama} perform significantly worse than vision-only open-source models, suggesting that the mere inclusion of audio is not sufficient. The limited gains from multimodal inputs can likely be attributed to two key factors:

\begin{itemize}[leftmargin=4.5mm, itemsep=0pt, topsep=0pt]
\item Imperfect Audio-Visual Alignment: Many models may not be well-trained on temporally synchronized audio and video inputs. Misaligned representations hinder the ability to leverage audio cues for grounding or disambiguation, especially in dynamic counting scenarios.

\item Insufficient Training on Grounding Tasks: The counting task inherently requires fine-grained spatio-temporal reasoning and object tracking, which are not emphasized in the pretraining of most MLLMs. The absence of dedicated supervision on such tasks limits their generalization to grounding-sensitive benchmarks.
\end{itemize}

In particular, the consistently low WCS scores across all models highlight a fundamental challenge: the task requires not only accurate grounding and precise counting, but also the ability to produce answers in a predefined format. This jointly stresses the models’ spatial understanding, numerical reasoning, and controllable output generation.

In summary, the counting task remains a difficult benchmark due to its dependence on precise grounding and multimodal reasoning. These results underscore the need for more targeted training strategies and benchmarks to foster genuine multimodal understanding rather than surface-level fusion.



\begin{table*}[t!]
\centering
\caption{\textbf{Performance of various MLLMs on CG-AV-Counting.} We use different colors to indicate the model categories: \colorbox[HTML]{d3f4ff}{Vision Language Models} and \colorbox[HTML]{ebdef0}{Audio-Visual Language Models}. \textbf{Bold} indicates the best scores, while \underline{underline} denotes the second-best scores.  }
\label{tab:benchmark}
\setlength\tabcolsep{2.5pt}
\renewcommand{\arraystretch}{1.05}
\resizebox{\textwidth}{!}{
\begin{tabular}{lc|cccc|cccc|cc}
\hline
\multirow{2}{*}{Model} & \multirow{2}{*}{Modality} & \multicolumn{4}{c|}{Black-box Evaluation (Long Acc)} & \multicolumn{4}{c|}{Black-box Evaluation (Ref Acc)} & \multicolumn{2}{c}{White-box Evaluation}\\
& & Acc$\uparrow$ & OBOA$\uparrow$ & MAE$\downarrow$ & RMSE$\downarrow$ & Acc$\uparrow$ & OBOA$\uparrow$ & MAE$\downarrow$ & RMSE$\downarrow$ & WCS$\uparrow$ & IFA$\uparrow$ \\
\midrule
Random & - &1.56 & 4.97 & 30.35 & 36.66  & - & - & - & - & 0.25 & 100.00 \\
Human (full-video) & A+V & 85.00 & 96.49 & 0.65 & 0.95 & 91.53 & 98.05 & 0.23 & 0.45 & 71.93 & 100.00 \\
GPT-4.1 (text)~\cite{openai_2025} & - & 6.04 & 13.15 & 8.90 & 17.60  & - & - & - & - & 0.00 & 76.24 \\
\midrule
\multicolumn{3}{l}{\textbf{\textit{Close-Source MLLMs}}} \\
\midrule
\rowcolor{lightblue} GPT-4.1~\cite{openai_2025} & V & 27.17 & 49.95 &  \underline{2.68} &  \textbf{4.78} & 37.39 & 64.85 & \underline{1.77} & \textbf{3.53} &  2.78 & \underline{98.73} \\
\rowcolor{lightblue} GPT-4o~\cite{hurst2024gpt} & V& 22.30 & 45.57 &  3.25 &  5.82 & 32.91 & 58.62 & 2.06 & 4.37  & 2.59 & \underline{98.73}\\
\rowcolor{lightblue} SEED-1.5 VL~\cite{guo2025seed1} & V & 27.85 & 52.19 &  2.93 &  6.39 & 36.12 & 57.64 & 2.35 & 4.45 & 2.46  & \textbf{99.03} \\

\rowcolor{lightpink}Gemini 2.5 Flash~\cite{team2023gemini} & A+V & \underline{36.90} & \underline{61.05} &  2.71 &  6.32 & \underline{41.48} & \underline{65.53} & 1.86 & \underline{4.08}  & \underline{4.20} &  95.03 \\
\rowcolor{lightpink}Gemini 2.5 Pro~\cite{team2023gemini} & A+V & \textbf{40.80} & \textbf{65.82} &  \textbf{2.33} &  7.20 & \textbf{47.42} & \textbf{72.25} & \textbf{1.47} & \textbf{3.53}  & \textbf{6.71} &  95.03 \\
\midrule
\multicolumn{3}{l}{\textbf{\textit{Open-Source MLLMs}}} \\
\midrule

\rowcolor{lightblue} VideoLLaMA3-7B~\cite{zhang2025videollama} & V & 12.46 & 30.57 &  4.52 &  12.60 & 18.50 & 41.48 & 3.23 & 5.68 & 1.03 & 91.72 \\
\rowcolor{lightblue} Eagle2-9B~\cite{li2025eagle} & V & 12.46 & 34.08 &  3.84 &  6.33 & 21.91 & 45.37 & 3.24 & 5.77 & 0.57  & 76.14 \\
\rowcolor{lightblue} Qwen2.5-VL-7B~\cite{bai2025qwen2} & V & 20.84 & 48.00 &  4.08 &  7.91 & 27.45 & 52.68 & 2.76 & 6.17 & 0.87 & 85.78 \\
\rowcolor{lightblue} MiniCPM-V 2.6~\cite{yao2024minicpm} & V & 13.83 & 33.20 & 4.06 & 6.58 & 18.99 & 39.44 & 3.62 & 6.74 & 0.57  & 67.67 \\
\rowcolor{lightblue} InternVL3-8B~\cite{zhu2025internvl3} & V & 17.92 & 43.43 &  3.21 &  5.55 & 30.09 & 56.67 & 2.57 & 5.76  & 0.71 & 97.57 \\ 
\rowcolor{lightblue} InternVideo2.5-8B~\cite{wang2025internvideo} & V & 22.20 & 48.30 &  3.17 & 5.92 & 28.33 & 56.47 & 2.75 & 5.70 & -  & - \\ 
\rowcolor{lightblue}
VideoChat-Flash-7B~\cite{li2024videochat} & V & 19.86 & 43.91 & 3.49 & 6.17 & 25.41 & 46.54 & 3.53 & 6.48 & -  & - \\
\rowcolor{lightblue} Eagle2.5-8B~\cite{chen2025eagle} & V & 28.04 & 51.80 &  2.76 &\underline{5.23}&  34.47 & 63.97 & 2.07 & 4.39  & 2.59  & 83.35 \\
\rowcolor{lightpink} UnifiedIO-2 XXL~\cite{lu2024unified} & A+V & 10.61 & 30.48 &  3.99 &  6.30 & 15.29 & 37.49 & 3.38 & 5.79 & 0.00 & 2.24 \\
\rowcolor{lightpink} VideoLLaMA2.1-7B-AV~\cite{cheng2024videollama} & A+V & 5.06 & 13.73 & 5.11 &  7.34 & 8.57 & 18.40 & 4.93 & 7.24 & 0.11 & 13.83 \\
\rowcolor{lightpink} Qwen2.5-Omni-7B~\cite{xu2025qwen2} & A+V & 22.30 & 48.69 &  3.92 &  8.49 & 33.05 & 60.20 & 3.05 & 5.79  & 1.17 & 95.52\\
\rowcolor{lightpink} Ola-7B~\cite{liu2025ola} & A+V & 17.92 & 38.85 &  4.57 &  10.52 & 25.33 & 46.53 & 3.30 & 5.98 & 0.84 & 75.66 \\

\hline
\end{tabular}
}

\end{table*}
\section{A Strong Baseline: AV-Reasoner}
\subsection{Overview}
\begin{figure}
  \centering
  \includegraphics[width=\linewidth]{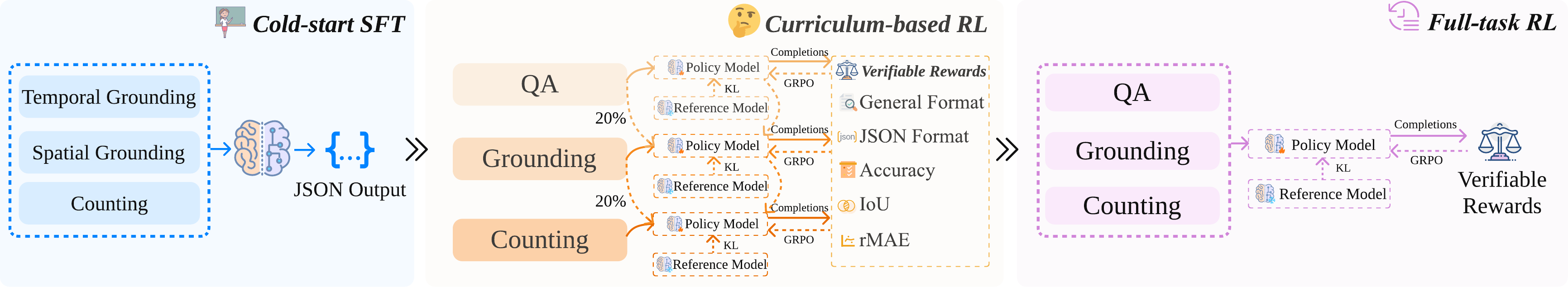}
  \caption{\textbf{Overview of Our Training Strategy.} Our strategy consists of three stages. During the curriculum-based RL stage, we introduce a Stage Review Mechanism, which mitigates forgetting of previously trained tasks by mixing in a proportion of earlier samples.}
  \label{fig: pipeline}
\end{figure}
Due to the lack of enough data for counting tasks, directly improving counting performance with existing data is suboptimal. To explore a way to improve models' counting performance without relying on large-scale annotated counting data, we introduce AV-Reasoner, built upon the Ola-Omni-7B\cite{liu2025ola} base model, which improves counting performance by training on tasks strongly related to counting, such as temporal and spatial grounding. This enables AV-Reasoner to enhance its counting ability through task-related exploration.

\subsection{Training Dataset}
\label{sec: dataset}
To tackle audio-visual counting tasks, the model needs core capabilities in audio-visual understanding, temporal grounding, and spatial grounding. As shown in Tab.~\ref{tab:datasets}, we use original annotations from datasets like AVQA~\cite{yang2022avqa}, MUSIC-AVQA~\cite{Li2022Learning}, AVE~\cite{tian2018ave}, UnAV~\cite{geng2023dense}, DVD-Counting~\cite{video-r1_2025}, and RepCount~\cite{hu2022transrac}. For the LLP~\cite{jatin2021cmlavvp} dataset, we use pseudo-labels from AVUIE~\cite{du2025crab}. For ARIG, we convert segmentation masks from AVSS~\cite{zhou2024avss} into bounding boxes. Unlike AVUIE~\cite{du2025crab}, which normalizes coordinates, our method retains the original image resolution. 

\begin{wraptable}{r}{0.5\textwidth} 
\centering
\small
\caption{The datasets used during training.}
\label{tab:datasets}
\begin{tabular}{ll}
\toprule
Tasks & Datasets \\
\midrule
AVQA & AVQA~\cite{yang2022avqa}, Music AVQA~\cite{Li2022Learning} \\
AVTG & AVE~\cite{tian2018ave}, UnAV~\cite{geng2023dense}, LLP~\cite{jatin2021cmlavvp}  \\
ARIG & AVSS-ARIG~\cite{zhou2024avss} \\
Counting & DVD-Counting~\cite{video-r1_2025}, RepCount~\cite{hu2022transrac} \\
\bottomrule
\end{tabular}
\end{wraptable}

\subsection{Verifiable Rewards Design}
We explore GRPO-based training for three major audio-visual tasks: QA, Grounding, and Counting. For each, we design a verifiable reward to guide learning.

\textbf{General Format Reward $R_{\text{GFormat}}$.} For QA and Counting tasks, we use the format reward function from DeepSeek-R1~\cite{guo2025deepseek} to ensure the model's output follows the expected format. The model must output thinking process within a \texttt{<think>...</think>} block and the final answer in a \texttt{<answer>...</answer>} block. We verify the format using regular expression matching. If correct, a reward of 1 is given; otherwise, 0.

\textbf{Json Format Reward $R_{\text{JFormat}}$.} For grounding tasks, we require the model to output in JSON format within the \texttt{<answer>...</answer>} block. Based on the general format reward, we introduce a JSON validity check. First, we attempt to parse the content between \texttt{<answer>} and \texttt{</answer>} as a standard JSON object. If parsing succeeds, the multiplier $m = 1.0$. If it fails, we use a bracket-matching algorithm to extract a potential JSON structure; if successful, $m = 0.5$. If both approaches fail, $m = 0$. Next, we define the key completeness score $s = N_{\text{complete}}/N_{\text{total}}$, where $N_{\text{complete}}$ is the number of JSON items with all required keys, and $N_{\text{total}}$ is the total number of items in the parsed list. The final JSON Format Reward $R_{\text{JFormat}} = m \times s$.

\textbf{Accuracy Reward $R_{\text{acc}}$.} For multiple-choice and closed-set question answering tasks, the reward is 1.0 for a correct prediction and 0.0 for an incorrect prediction.

\textbf{IoU Reward $R_{\text{IoU}}$.} For object and event grounding tasks, we calculate the tIoU or cIoU~\cite{zheng2021enhancing} between predicted and ground truth sets of temporal segments or spatial bounding boxes. The $R_{\text{IoU}}$ is the average of the IoU scores across all groups, where the best matches are found using a greedy algorithm for optimal alignment. If both the prediction and reference sets are empty, the reward is set to 1.0.

\textbf{rMAE Reward $R_{\text{rMAE}}$.} For counting tasks, we use relative MAE (Mean Absolute Error) to compute the reward. The calculation method is as follows:
\begin{align}
R_{\text{rMAE}} = 1 - \min\left(1.0, {|\text{Pred} - \text{GT}|}/{\text{GT}}\right)
\end{align}
Specially, if the ground truth is 0, we use accuracy-based $R_{\text{acc}}$ as the $R_{\text{rMAE}}$.

\subsection{Training Strategy}
As shown in Fig.~\ref{fig: pipeline}, our training pipeline follows a three-stage process: cold-start SFT, curriculum-based RL with stage review mechanism, and full-task RL.

\textbf{Cold-start SFT.} To improve Ola-Omni's~\cite{liu2025ola} performance on counting and grounding tasks, we conduct a cold-start SFT, training it on AVTG, ARIG, and counting tasks. This phase also enhances its ability to generate structured JSON outputs for rule-based rewards in these tasks.

\textbf{Curriculum-based RL (CB-RL) with Stage Review Mechanism.} Training all tasks simultaneously from scratch can negatively impact performance on complex tasks like counting, which suffers from limited data availability. To mitigate this, we use a curriculum learning strategy, organizing tasks into three levels of increasing difficulty: (1) question answering, (2) temporal and spatial grounding, and (3) counting. The model is trained on each level for two epochs in sequence using GRPO, allowing it to progressively acquire necessary skills. To improve training efficiency, we introduce an offline data filtering pipeline: before each epoch, we perform five rollouts per sample using the reference model. For QA and counting tasks, we discard samples where all rollouts yield correct answers; for grounding tasks, we discard those with the average IoU greater than 0.9. 

Additionally, to address performance degradation caused by forgetting earlier tasks during later training stages, we introduce a \textbf{Stage Review Mechanism (SRM)} that mixes in 20\% of previously seen samples to maintain stability across tasks.

\textbf{Full-task RL  (FT-RL).} Despite using the SRM, some performance degradation across tasks remains. To address this, we conduct a full-task training stage after curriculum learning. We sample data from each dataset, ensuring balanced distribution across tasks and difficulty levels, with difficulty defined by the pass rate over five rollouts. Using these samples, we perform another round of reinforcement learning to enhance the model's overall performance.
\section{Expirements}
\subsection{Comparison with State-of-the-Art MLLMs}

\begin{table*}[t!]
\centering
\caption{\textbf{Comparison with SoTA MLLMs on Various Audio-Visual Benchmarks}. We use different colors to indicate the model categories: \colorbox[HTML]{d3f4ff}{Vision Language Models}, \colorbox[HTML]{ebdef0}{existing Audio-Visual Language Models} and \colorbox[HTML]{ffe9b8}{our proposed model}.}
\label{tab:compare_sota_video}
\setlength\tabcolsep{2.5pt}
\renewcommand{\arraystretch}{1.18}
\resizebox{\textwidth}{!}{
\begin{tabular}{l|c|cc|c|cc|c|ccc|cc|c|c|c}
\hline
\multirow{2}{*}{Model}  & MusicAVQA  & \multicolumn{2}{c|}{LLP} & UnAV & \multicolumn{2}{c|}{ARIG} & DVD-Counting &\multicolumn{3}{c|}{CG-AV-Counting}& \multicolumn{2}{c|}{AVHBench} & AV-Odyssey & OmniBench & WorldSense\\
                       & acc & Segment Level & Event Level & mAP & cIoU & AUC & acc & long & ref & WCS  & A2V & V2A & acc & acc & acc \\
\midrule
\multicolumn{3}{l}{\textbf{\textit{Close-Source MLLMs}}} \\
\midrule
\rowcolor{lightblue} GPT-4.1~\cite{openai_2025} &  - &   - & - & - & - & - & - &27.17&37.39&2.78& - & - & - & - & - \\
\rowcolor{lightblue} GPT-4o~\cite{hurst2024gpt} &  - &   - & - & - & - & - & - &22.30&32.91&2.59& - & - & \underline{34.50} & 51.14 & 42.60 \\
\rowcolor{lightblue} Claude 3.5 Sonnet~\cite{anthropic}& - &  - & - & - & - & - & - & - &-&-& -&- & - & \textbf{59.37} & 34.8 \\
\rowcolor{lightpink} Gemini 2.5 Pro~\cite{team2023gemini} &  - &   - & - & - & - & - & - &\textbf{40.80}&\textbf{47.42}&\textbf{6.71}& - & - & - & - & - \\
\rowcolor{lightpink} Gemini 2.5 Flash~\cite{team2023gemini} &  - &   - & - & - & - & - & - &\underline{36.90}&\underline{41.48}&\underline{4.20}& - & - & - & - & - \\
\rowcolor{lightpink} Gemini 1.5 Pro~\cite{team2023gemini} & - &  - & - & - & - & - & - & -&-&-&- & - & 30.80 & 42.91 & \textbf{48.00} \\
\rowcolor{lightpink} Gemini 1.5 Flash~\cite{team2023gemini} &  - &  - & - & - & - & - &-&-&-& - & \underline{83.30} & 63.00 & 27.80 & - & - \\

\midrule
\multicolumn{3}{l}{\textbf{\textit{Open-Source MLLMs}}} \\
\midrule
\rowcolor{lightblue} Video-R1~\cite{feng2025video} &  - &  - & - & - & - & - & 34.50 & -&-&- & -&- & - & - & - \\
\rowcolor{lightpink} Qwen2.5-Omni~\cite{xu2025qwen2}  &  - &  - & - & - & - & - & - &22.30&33.05&1.17& -&-&- & \underline{56.13} & 45.40 \\
\rowcolor{lightpink} UnifiedIO2-XXL~\cite{lu2024unified}  &  - &  - & - & - & - & - & - & 10.61&15.29&0.00&- & - & 27.20 & 33.98 & 25.90 \\
\rowcolor{lightpink} VideoLLaMA 2~\cite{cheng2024videollama}  &  - &  - & - & - & - & - & - & 5.06&8.57&0.11&75.20 & 74.20 & 26.80 & - & 25.40
 \\
\rowcolor{lightpink} Video-SALMONN~\cite{sun2024video} &  47.60  &  - & - & - & - & - & -&-&-&- & 78.10 & 65.20 & - & 35.64 & - \\
\rowcolor{lightpink} OneLLM~\cite{han2024onellm} & 47.60  &  - & - & - & - & - & -&-&- &-& 53.70 & 44.30 & 27.40 & - & 22.80 \\
\rowcolor{lightpink} GroundingGPT~\cite{li2024groundinggpt} &  - &  - & - & - & 44.02 & 0.45 &-&-&-&- & - & - & - & - & - \\
\rowcolor{lightpink} VALOR~\cite{liu2024valor} & 78.90 &  - & - & - & - & - & - & - & -&-&-&- & - & - & - \\
\rowcolor{lightpink} X-InstructBLIP~\cite{panagopoulou2024x} & 44.50  &  - & - & - & - & -&-&- &-& - & 18.10 & 16.30 & - & - & - \\
\rowcolor{lightpink} MEERKAT~\cite{chowdhury2024meerkat} &  79.15  &  54.96 & - & - & - & -&-&- &-& - & - & - & - & - & - \\
\rowcolor{lightpink} AVicuna~\cite{tang2024avicuna} &  49.60  &  - & - & 60.30 & - & - & -&-&- & -&- & - & - & - & - \\
\rowcolor{lightpink} PAVE~\cite{liu2025pavepatchingadaptingvideo} &  82.30  &  - & - & - &-&-& - &-& - & - & - & - & - & - & - \\
\rowcolor{lightpink} Crab~\cite{du2025crab} &  78.94  &  59.00 & 54.44 & - & 41.78 & 0.42 & -&-&-&- & - & - & - & - & - \\
\rowcolor{front-color} 
AV-Reasoner (Ours)  & \underline{83.39} & \underline{62.40} & \underline{59.40} & \underline{64.69} & \underline{45.06} & \underline{0.47} & \underline{43.50} & 22.30&35.83&1.11&\textbf{84.45} & \textbf{80.63} & \textbf{36.38} & 48.95 & \underline{45.84} \\
\rowcolor{front-color} 
AV-Reasoner-Thinking (Ours)  & \textbf{85.01} & \textbf{64.70} & \textbf{62.00} & \textbf{65.18} & \textbf{46.73} & \textbf{0.49} & \textbf{44.00} &21.03&34.08&1.68& 82.45 & \underline{80.28} & 34.29 & 48.34 & 44.36 \\
\hline
\end{tabular}}
\end{table*}

As shown in Tab.~\ref{tab:compare_sota_video}, our model, AV-Reasoner, demonstrates strong performance across various audio-visual tasks. AV-Reasoner-Thinking achieves an accuracy of 85.01 on MusicAVQA~\cite{Li2022Learning}, surpassing PAVE's 82.30. For event localization, it achieves 64.70 (segment-level) and 62.00 (event-level) on LLP~\cite{jatin2021cmlavvp}, and 65.18 mAP on UnAV~\cite{geng2023dense}, all significantly surpassing the current SOTA results. In spatial grounding, AV-Reasoner-Thinking outperforms GroundingGPT~\cite{li2024groundinggpt} and Crab~\cite{du2025crab}, with 46.73 cIoU and 0.49 AUC on ARIG~\cite{zhou2024avss}. On DVD-Counting~\cite{video-r1_2025}, AV-Reasoner achieves 44.00 accuracy, surpassing Video-R1~\cite{feng2025video} by 9.50 points.

Beyond task-specific benchmarks, AV-Reasoner-Thinking also excels in comprehensive benchmarks. In AV-Odyssey~\cite{gong2024av}, it reaches 34.29 accuracy, ranking just below GPT-4o~\cite{hurst2024gpt}. When not forced to output thinking, AV-Reasoner surpasses GPT-4o by 1.88 points. On WorldSense~\cite{hong2025worldsense}, AV-Reasoner scores 45.84 without thinking output, second only to Gemini 1.5 Pro~\cite{team2023gemini}, while dropping to 44.36 with thinking output. In OmniBench~\cite{li2024omnibench}, AV-Reasoner scores 48.95, though it still lags behind Qwen2.5-Omni~\cite{xu2025qwen2}, it outperforms Gemini 1.5 Pro~\cite{team2023gemini}. The performance difference between explicitly outputting the thinking process and not will be discussed in the ablation study section.
\subsection{Comparison with Base Model on Counting Tasks}

\begin{table*}[t!]
\centering
\caption{\textbf{Comparison with Base Model on Counting Tasks.} Models trained with our proposed strategy consistently achieve higher accuracy across all counting benchmarks. The setting without explicit reasoning outputs yields even better performance.}
\label{tab:compare_counting}
\setlength\tabcolsep{2.5pt}
\renewcommand{\arraystretch}{1.3}
\resizebox{\textwidth}{!}{
\begin{tabular}{l|cc|ccc|cccc|cccc|cc}
\hline
\multirow{2}{*}{Model}  & \multicolumn{2}{c|}{DVD-Counting}  & \multicolumn{3}{c|}{WorldSense} & \multicolumn{4}{c|}{CG-AV-Counting (Long Acc)} & \multicolumn{4}{c|}{CG-AV-Counting (Ref Acc)} & \multicolumn{2}{c}{CG-AV-Counting (White Box)}\\
                       & Acc & OBOA & Object & Action & Audio & Acc & OBOA & MAE & RMSE & Acc & OBOA & MAE & RMSE & WCS & IFA \\
\hline
Ola-Omni (Base Model) & 16.50 & 51.50 & 36.10 & 30.91 & 44.44 & 17.92 & 38.85 & 4.57 & 10.52 & 25.33 & 46.53 & 3.30 & 5.98 & 0.84 & 75.66 \\
\hline
\multirow{2}{*}{AV-Reasoner-Thinking (Ours)} & 44.00 & 81.00 & 36.59 & 34.55 & 46.67 & 21.03 & 48.78 & 3.26 & 8.20 & 34.08 & 60.95 & 2.40 & 4.66 & 1.68 & 79.65 \\
                      & \textcolor{red}{(+27.50)} & \textcolor{red}{(+29.50)} & \textcolor{red}{(+0.49)} & \textcolor{red}{(+3.64)} & \textcolor{red}{(+2.23)} & \textcolor{red}{(+3.11)} & \textcolor{red}{(+9.93)} & \textcolor{red}{(-1.31)} & \textcolor{red}{(-2.32)} & \textcolor{red}{(+8.75)} & \textcolor{red}{(+14.42)} & \textcolor{red}{(-0.90)} & \textcolor{red}{(-1.32)} & \textcolor{red}{(+0.84)} & \textcolor{red}{(+3.99)} \\
\hline
\multirow{2}{*}{AV-Reasoner (Ours)} & 43.50 & 80.50 & 37.07 & 32.12 & 47.78 & 22.30 & 48.30 & 3.15 & 5.89 & 35.83 & 61.44 & 2.38 & 4.44 & 1.11 & 80.82 \\
             & \textcolor{red}{(+27.00)} & \textcolor{red}{(+29.00)} & \textcolor{red}{(+0.97)} & \textcolor{red}{(+1.21)} & \textcolor{red}{(+3.34)} & \textcolor{red}{(+4.38)} & \textcolor{red}{(+9.45)} & \textcolor{red}{(-1.42)} & \textcolor{red}{(-4.63)} & \textcolor{red}{(+10.50)} & \textcolor{red}{(+14.91)} & \textcolor{red}{(-0.92)} & \textcolor{red}{(-1.54)} & \textcolor{red}{(+0.27)} & \textcolor{red}{(+5.16)} \\
\hline
\end{tabular}}
\end{table*}

Tab.~\ref{tab:compare_counting} highlights the effectiveness of GRPO in improving counting capabilities. Compared to the base model Ola-Omni~\cite{liu2025ola}, both AV-Reasoner variants achieve consistent improvements across all counting benchmarks. This demonstrates that our GRPO-based training strategy not only leverages existing grounding and QA abilities, but also enables the model to generalize them to counting.

\subsection{Ablation Studies}
\begin{figure}

  \centering
  \includegraphics[width=\linewidth]{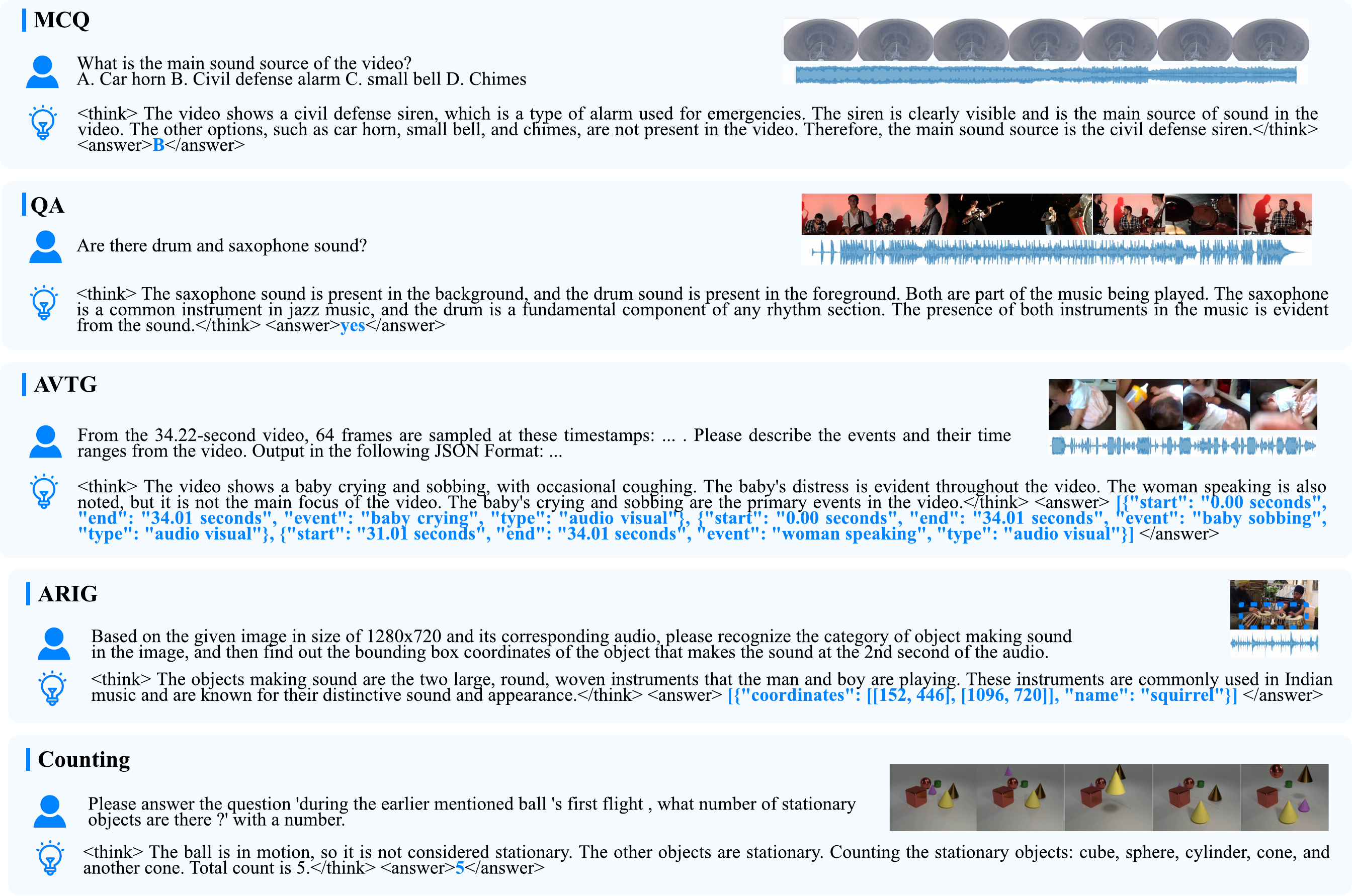}
  \caption{\textbf{Example Ouputs on Various Tasks.} AV-Reasoner not noly outputs the correct answer, but also give the thinking process.}
  \label{fig: example}
\end{figure}

In this section, we aim to address the following key questions through a series of experiments:

\textbf{Q1: How does each stage of our training strategy enhance the model's counting capability?} 

As shown in Tab.~\ref{tab:rft}, SFT improves in-domain DVD-Counting~\cite{video-r1_2025} accuracy but suffers from overfitting, leading to degraded out-of-domain performance. QA-based RFT mitigates overfitting and improves generalization, but its gains on counting are limited. In contrast, grounding-based training greatly improves counting by focusing on the skills needed for it. And counting-specific and full-task RL further improve performance by reinforcing task-relevant abilities.

\begin{table*}[t!]
\centering

\caption{Impact of Each Training Stage on Counting Performance. SRM is applied to CB-RL stages.}
\label{tab:rft}
\setlength\tabcolsep{2.5pt}
\renewcommand{\arraystretch}{1.1}
\resizebox{\textwidth}{!}{
\begin{tabular}{l|cccc|ccc|cccc|cccc}
\hline

\multirow{2}{*}{Stage} & \multicolumn{4}{c|}{DVD-Counting} & \multicolumn{3}{c|}{WorldSense} & \multicolumn{4}{c|}{CG-AV-Counting(Long Acc)} & \multicolumn{4}{c}{CG-AV-Counting(Ref Acc)}\\
                       & Acc & OBOA & MAE & RMSE & Object & Action & Audio & Acc & OBOA & MAE & RMSE & Acc & OBOA & MAE & RMSE\\
\midrule
Base Model&16.50 & 51.50 & 1.86 & 2.45 & 36.10 & 30.91 & 44.44 & 17.92 & 38.85 &4.57 &10.52 & 25.33 & 46.53 &3.30 & 5.98\\
SFT & 41.50 & 75.00 & 0.99 & 1.51 & 27.80 & 30.91 & 35.52 & 15.00 & 35.25 &3.79 &8.24 & 23.89 & 44.34 &2.91 & 4.68 \\
SFT + $\text{CB-RL}_{\text{QA}}$ & 23.00 & 54.00 & 1.77 & 2.44 & 33.66 & 31.52 & 41.46 & 16.55 & 37.39 &4.75 &8.27 & 24.76 & 45.68 &2.73 & 4.68 \\
SFT + $\text{CB-RL}_{\text{QA}}$ + $\text{CB-RL}_{\text{Grounding}}$ & 34.50 & 70.00 & 1.32 & 1.96 & 34.63 & 33.17 & 45.37 & 18.21 & 45.37 &4.07 &8.26 & 28.40 & 51.04 &2.54 & \underline{4.67} \\
SFT + $\text{CB-RL}_{\text{QA}}$ + $\text{CB-RL}_{\text{Grounding}}$ + $\text{CB-RL}_{\text{Counting}}$ & \underline{43.00} & \underline{80.00} & \underline{0.85} & \textbf{1.28} & \underline{35.12} & \underline{33.66} & \underline{45.85} & \underline{20.84} & \underline{46.54} &\underline{3.31} &\underline{8.22} & \underline{34.00} & \underline{59.60} &\underline{2.50} & 4.68\\
SFT + $\text{CB-RL}_{\text{QA}}$ + $\text{CB-RL}_{\text{Grounding}}$ + $\text{CB-RL}_{\text{Counting}}$ + \text{FT-RL} & \textbf{44.00} & \textbf{81.00} & \textbf{0.84} & \underline{1.29} & \textbf{36.59} & \textbf{34.55} & \textbf{46.67} & \textbf{21.03} & \textbf{48.78} &\textbf{3.26} &\textbf{8.20} & \textbf{34.08} & \textbf{60.95} &\textbf{2.40} & \textbf{4.66} \\
\hline
\end{tabular}}
\end{table*}

\textbf{Q2: What is the impact of curriculum-based RL and the review mechanism?} 

As shown in Tab.~\ref{tab:cb}, the base model performs poorly on grounding tasks due to its weak ability to follow JSON-structured output formats. SFT mitigates this issue by explicitly guiding the model to produce well-formed outputs, which is critical for grounding evaluation. While directly applying RL without SFT can improve performance on simpler tasks like QA, it brings limited benefits on more complex tasks that demand reasoning and precise grounding.

Training with GRPO on all tasks at once highlights the model's poor generalization under limited data, especially for counting tasks. CB-RL helps by learning tasks of varying difficulty progressively, but degrades performance on previously learned tasks. Our SRM addresses this by periodically revisiting earlier data. When combined with FT-RL, it achieves the best trade-off between learning new capabilities and preserving prior knowledge.

\begin{table*}[t!]
\centering
\caption{Effect of Curriculum Learning and Review Mechanism on Model Performance.}
\label{tab:cb}
\setlength\tabcolsep{2.5pt}
\renewcommand{\arraystretch}{1.1}
\resizebox{\textwidth}{!}{
\begin{tabular}{l|cc|ccccc|ccccccc}
\hline

\multirow{2}{*}{Method} & \multicolumn{2}{c|}{QA} & \multicolumn{5}{c|}{Grounding} & \multicolumn{7}{c}{Counting} \\
                       &AVQA & MUSIC-AVQA & AVE &$\text{LLP}_{\text{segment}}$ & $\text{LLP}_{\text{event}}$ & UnAV & ARIG & DVD & $\text{CG-AV}_{\text{long}}$ & $\text{CG-AV}_{\text{ref}}$ & $\text{CG-AV}_{\text{WCS}}$ & $\text{WorldSense}_{\text{obj}}$ & $\text{WorldSense}_{\text{act}}$ & $\text{WorldSense}_{\text{aud}}$ \\
\midrule
Base model & 82.61 & 79.47 & 19.43 & 14.5 & 14.5 &0.28& 0.19 & 16.50 & 17.92 & 25.33 & 0.84 & \underline{36.10} & 30.91 & 44.44  \\
\text{CB-RL}(w/ SRM) + FT-RL& 92.97 & 84.59 & 78.56 & 51.90 & 49.80 &55.45& 45.55 & 29.50 & 18.31 & 28.41 & 1.15 & \textbf{36.59} & 31.71 & 45.37  \\
SFT & 78.73 & 78.10 & 80.02 & 48.70 & 47.10 &52.67& 45.48 & 41.50 & 15.00 & 23.89 & 0.83 & 27.80 & 30.91 & 35.52  \\
SFT + GRPO with all data &\textbf{93.25} & \textbf{85.68} & 79.82 & 58.10 & 55.80 & 60.09 & \textbf{48.43} & 31.50 & 10.42 &17.16 &0.77 & 24.39 & 22.42 & 32.20\\
SFT + \text{CB-RL}(w/o SRM) &89.24 & 78.93 & 80.03 & 60.30 & 58.30  & 63.91 & 45.97 & 42.50 & \underline{20.84}  &33.07  & 1.19 & 34.63 &30.73 & 42.44\\
SFT + \text{CB-RL}(w/o SRM) + FT-RL &93.13 & 83.44 & 80.10 & 61.30 & 60.90 & 64.28 & 46.72 & \underline{43.50} & 20.74 &33.68 &1.61 & 35.61 & 33.17 & \underline{46.34}\\
SFT + \text{CB-RL}(w/ SRM) &93.14 & 84.79 & \underline{80.15} & \underline{64.50} & \underline{61.60} & \underline{65.07} & 46.62 & 43.00 & \underline{20.84} &\underline{34.00} &\underline{1.66} & 35.12 &\underline{33.66} & 45.85\\
SFT + \text{CB-RL}(w/ SRM) + FT-RL &\underline{93.17} & \underline{85.01} & \textbf{81.26} & \textbf{64.70} & \textbf{62.00} & \textbf{65.18} & \underline{46.73} & \textbf{44.00} & \textbf{21.03} &\textbf{34.08} &\textbf{1.68} & \textbf{36.59} &\textbf{34.55} & \textbf{46.67}\\
\hline
\end{tabular}}
\end{table*}

\textbf{Q3: Is explicit thinking output truly beneficial for improving model performance?}

\begin{wraptable}{r}{0.5\textwidth} 
\vspace{-4mm}
\centering
\small
\caption{Comparison of Hallucination on Benchmark Subset with and without Explicit Thinking.}
\label{tab:thinking}
\begin{tabular}{lcc}
\toprule
Benchmark & Thinking & No Thinking \\
\midrule
AVHBench A2V  & 82.45 &  84.45\\
AVHBench V2A & 80.28 & 80.63  \\
WorldSense Hall. & 35.56 & 45.56 \\
AVOdyssey Hall. & 31.00 &  32.50\\
\bottomrule
\end{tabular}
\vspace{-3mm}
\end{wraptable}

As shown in Tab.~\ref{tab:compare_sota_video}, GRPO enhances reasoning ability, but explicitly output thinking at inference does not always improve performance, which may introduce errors and increase hallucinations, particularly in out-of-domain settings. This is evident in benchmarks like WorldSense and AVOdyssey hallucination subsets, and AVHBench, where explicit thinking lowers accuracy. Specifically, as shown in Tab.~\ref{tab:thinking}, AVHBench audio-driven video hallucination accuracy drops from 84.45 to 82.45, and WorldSense Hallucination subset drops from 45.56 to 35.56. These findings emphasize the importance of balancing reasoning transparency with robustness to reduce hallucinations.

\subsection{Qualitative Results}
Fig.~\ref{fig: example} visualizes AV-Reasoner's outputs on tasks including MCQ, QA, AVTG, ARIG, and Counting. The model not only provides correct answers but also generates coherent reasoning. For example, in AVTG, it identifies the sounds and determines the primary one, and in ARIG, it explains how the sounding object is located based on shape and auditory cues.
\section{Conclusion}
\label{sec: conclusion}
In this paper, we introduce CG-AV-Counting, a manually-annotated multimodal benchmark for evaluating counting in long videos with clue annotations. It reveals that current MLLMs struggle in both black-box and white-box settings, particularly in reasoning counting. In order to explore ways to improve model's counting performance, we propose a GRPO-based training strategy. Our model AV-Reasoner achieves strong performance in counting and other audio-visual understanding tasks. 

\textbf{Limitation.} Firstly, white-box evaluation of object and attribute counting typically depends on predefined clue frames due to sparse clue annotations. Moreover, inconsistencies between the model's reasoning process and its final answers often lead to performance drops when the model is required to explicitly output its reasoning at the interface, highlighting the need to enforce reasoning-answer consistency during training.

\bibliography{sample}

\newpage
\appendix

\section{Benchmark Construction}
\subsection{Reference-Query Modality Definition}
To support comprehensive evaluation of multimodal counting capabilities, our benchmark defines five distinct modal settings based on the combination of reference and query modalities. Each setting reflects a different reasoning requirement, depending on which modality is used to locate the counting target and query interval:
\begin{itemize}[leftmargin=4.5mm]

\item \textbf{Visual-only}: The model is required to both locate and count using only visual input. This setting targets scenarios where the counting question and answer are entirely grounded in visual content.

\item \textbf{Audio-only}: The model must rely solely on audio cues to identify the relevant segments and determine the count. This setting emphasizes sound-based reasoning, such as counting distinct audio events or speaker turns.

\item \textbf{Visual-reference, Audio-query}: The model uses visual input to locate the relevant temporal segment and then performs counting based on audio information within that scope (e.g., “How many people spoke in the scene showing the conference table?”).

\item \textbf{Audio-reference, Visual-query}: The model leverages audio to identify the query interval and subsequently counts visually observable targets in the corresponding segments (e.g., “How many people are visible when the sound of clapping occurs?”).

\item \textbf{Joint Audio-Visual}: Both audio and visual modalities are required to solve the task effectively. The model must integrate information across modalities to interpret the question, localize relevant segments, and produce an accurate count. This setting reflects complex scenarios where neither modality alone is sufficient, or where combining both provides more reliable counting signals.

\end{itemize}

\subsection{Counting Target Definition}
We define three types of counting targets in our benchmark to evaluate model capabilities from multiple perspectives:
\begin{itemize}[leftmargin=4.5mm]
\item  \textbf{Event}: A temporally localized activity or incident in the video. Event counting requires the model to recognize distinct occurrences over time.

\item \textbf{Object}: A visually identifiable entity in the scene. Object counting involves detecting and enumerating instances. The model must avoid double-counting due to camera motion, scene transitions, or repeated appearances of the same entity.

\item \textbf{Attribute}: A clustering of objects based on shared visual or semantic properties. Rather than counting individual instances, the model must identify groupings according to the query attribute (e.g., “How many different clothing colors are worn by people?”).
\end{itemize}

\subsection{More Dataset Statistics}
\begin{figure}
  \centering
  \includegraphics[width=0.5\linewidth]{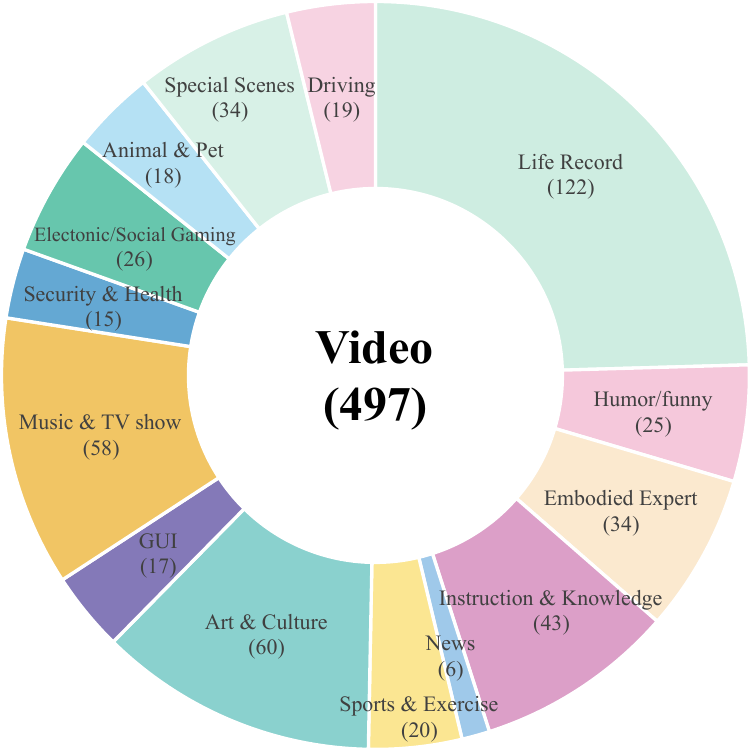}
  \caption{Statistics of Video Content Categories.}
  \label{fig:category}
\end{figure}
\begin{figure}
  \centering
  \includegraphics[width=0.5\linewidth]{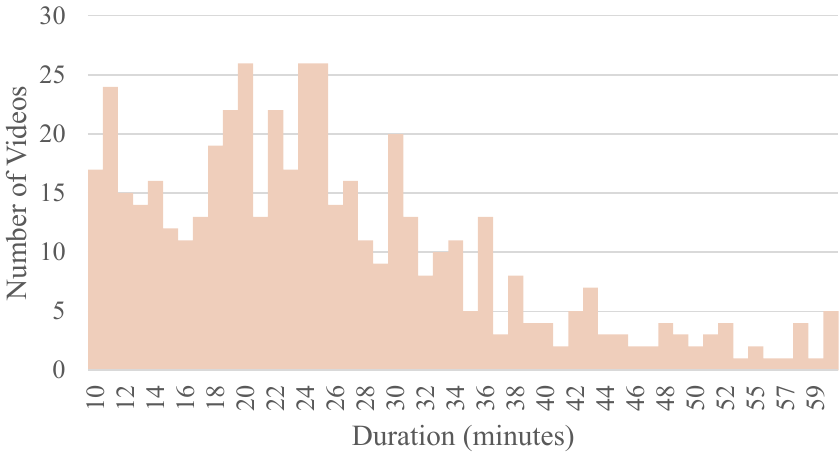}
  \caption{Statistics of Video Duration.}
  \label{fig:duration}
\end{figure}

As shown in Fig.~\ref{fig:category} and Fig.~\ref{fig:duration}, the videos selected from CG-Bench cover a diverse range of topics, including Life Record, Sports, Instruction, and TV Show, among more than ten categories. All videos are longer than 10 minutes, with the majority ranging from 20 to 30 minutes. This design encourages models to perform counting over long temporal contexts, which is essential for evaluating their capability in long-range temporal grounding and accumulation-based reasoning. This addresses a key limitation of existing counting benchmarks, which often rely on short clips that fail to capture the complexity of long-range dependencies.


\subsection{Prompt for Generating Initial Question Proposals}

\begin{lstlisting}
You are given a video, and your task is to generate **count-based audio-visual reasoning questions** that can be answered by analyzing a clearly defined segment of the video.

### **Definition of Count-Based Audio-Visual Reasoning Question**
Each question should involve counting something that is either **audibly heard**, **visually seen**, or both, during a localized event in the video. The question must be specific, grounded in real scenes, and the answer must be **objectively verifiable** within a given time span.

### **Modalities and Reasoning Types**
- **A2V (Audio to Visual)**: The question is triggered by an **audio cue** (e.g., a sound, noise, or dialogue) which helps the model **identify the relevant time segment** in the video. Once the time segment is located using the audio cue, the question asks about **visual information** (e.g., objects, people, or actions) within that segment. For example, "When the dog starts barking, how many people are visible in the background?" The audio cue identifies the time segment (e.g., when the dog barks), and the question asks about what is seen visually during that segment.

- **V2A (Visual to Audio)**: The question is triggered by a **visual cue** (e.g., someone entering the frame, a person performing an action) which helps the model **identify the relevant time segment** in the video. The question then asks about **auditory events** (e.g., how many sounds or spoken words are heard) that occur during that time segment. For example, "When the firefighter enters the building, how many sirens can be heard in the background?" Here, the visual cue (the firefighter entering) locates the time segment, and the question asks about sounds during that segment.

- **AV (Audio + Visual)**: The question requires both **audio and visual cues** to accurately count objects/events. Both cues help locate the relevant time segment in the video, and then the question asks for a count based on the interaction between both modalities. For example, "When the presenter gestures to the audience, how many people respond verbally?" Both audio (the people speaking) and visual (the presenter's gesture) cues are needed to define the time segment for counting.

- **A (Audio Only)**: The question is based solely on **audio cues** to identify a relevant time segment, and the task is to count auditory events within that segment. For example, "How many door slams can be heard in the scene?" The audio cue (the door slams) defines the time segment, and the model counts the auditory events within that period.

- **V (Visual Only)**: The question is based solely on **visual cues**, with no need for audio. The task involves counting visible objects or actions in the defined segment. For example, "How many people are wearing blue shirts in the crowd at the park?" This is based purely on visual observation.

### **Constraints**
1. Each question must correspond to a **specific and bounded video segment**.
2. The **answer must be a count** (e.g., number of visual objects, number of auditory events, number of multimodal occurrences, etc.).
3. The answer must be **clearly determined** and not ambiguous within the video span.
4. The difficulty of the question should be specified based on perceptual complexity (e.g., occlusion, background noise, overlapping motion/sound).
5. Include a **diverse mix** of A2V, V2A, AV, A, and V questions in the output.
6. Avoid redundant or trivial questions.
7. The question should not include specific timepoints.

### **Output Format**
Return your result as a **JSON array**, where each entry is a dictionary with the following fields:

- `"question"`: A clear and specific counting question.
- `"type"`: One of `"A2V"`, `"V2A"`, `"AV"`, `"A"`, or `"V"`.
- `"start_time"`: Start of relevant video segment (`"MM:SS"`).
- `"end_time"`: End of relevant video segment (`"MM:SS"`).
- `"counting_result"`: The correct count answer.

### **Example Output**
Enclose the JSON block within `<json></json>` tags.

<json>
[
  {
    "question": "When the dog starts barking, how many people are visible in the background?",
    "type": "A2V",
    "start_time": "01:15",
    "end_time": "01:45",
    "counting_result": 3
  },
  {
    "question": "When the firefighter enters the building, how many sirens can be heard in the background?",
    "type": "V2A",
    "start_time": "02:00",
    "end_time": "02:30",
    "counting_result": 2
  },
  {
    "question": "When the presenter gestures to the audience, how many people respond verbally?",
    "type": "AV",
    "start_time": "03:00",
    "end_time": "03:40",
    "counting_result": 4
  },
  {
    "question": "How many door slams can be heard in the scene?",
    "type": "A",
    "start_time": "05:15",
    "end_time": "05:45",
    "counting_result": 3
  },
  {
    "question": "How many people are wearing different colors of clothes in the park?",
    "type": "V",
    "start_time": "02:00",
    "end_time": "02:30",
    "counting_result": 4
  }
]
</json>

\end{lstlisting}

\subsection{Evaluation Prompts}
\subsubsection{Black-Box Evaluation}
\begin{lstlisting}
Watch the video and  answer the question '{Question Here}' with a number. Just output the number itself, don't output anything else.
\end{lstlisting}
\subsubsection{White-Box Evaluation (Event)}
\begin{lstlisting}
Watch the video and provide your answer to the question '{Question Here}', including the start and end timestamps for each event. Format your answer in JSON, enclosed in <answer> and </answer> tags. The output should look like this: <answer>[[\"start_time\", \"end_time\"], ...]</answer>. Ensure each timestamp is in seconds (e.g., 'xx.xx').
\end{lstlisting}
\subsubsection{White-Box Evaluation (Object)}
\begin{lstlisting}
According to the given video frames, answer the question '{Question Here}', including the bounding box for the query object in the first frame where it appears. For subsequent frames where the object appears, do not provide the bounding box again. Format your answer in JSON, enclosed within <answer> and </answer> tags. The output should look like this: <answer>{{\"Frame1\": [[x_min, y_min, x_max, y_max]], \"Frame2\": [...],...}}</answer>. In the output, each frame should either contain the bounding box of the object (if it appears for the first time in that frame) or an empty list `[]` (if the object does not appear or it has already been labeled in a previous frame). Ensure that bounding boxes are listed as [x_min, y_min, x_max, y_max].
\end{lstlisting}
\subsubsection{White-Box Evaluation (Attribute)}
\begin{lstlisting}
According to the given video frames, answer the question '{Question Here}', clustering the objects based on the question. For each unique cluster, assign a unique label and return the bounding box for each object in the first frame where it appears. For subsequent frames where the object appears, do not output anything. Format your answer in JSON, enclosed within <answer> and </answer> tags. The output should look like this: <answer>{{\"Frame 1\": [{{\"bbox\": [x_min, y_min, x_max, y_max], 'label': \"Label 1\"}}], \"Frame 2\": [...], ...}}</answer>. In the output, each frame should either contain the bounding box and label for the object (if it appears for the first time in that frame) or an empty list `[]` (if the object has already been labeled or does not appear in that frame). The label should correspond to a unique object cluster according to the question.
\end{lstlisting}

\subsection{More Evaluations}
\subsubsection{Model Performance across Different Counting Targets}

\begin{figure}[t]
    \centering
    \includegraphics[width=\linewidth]{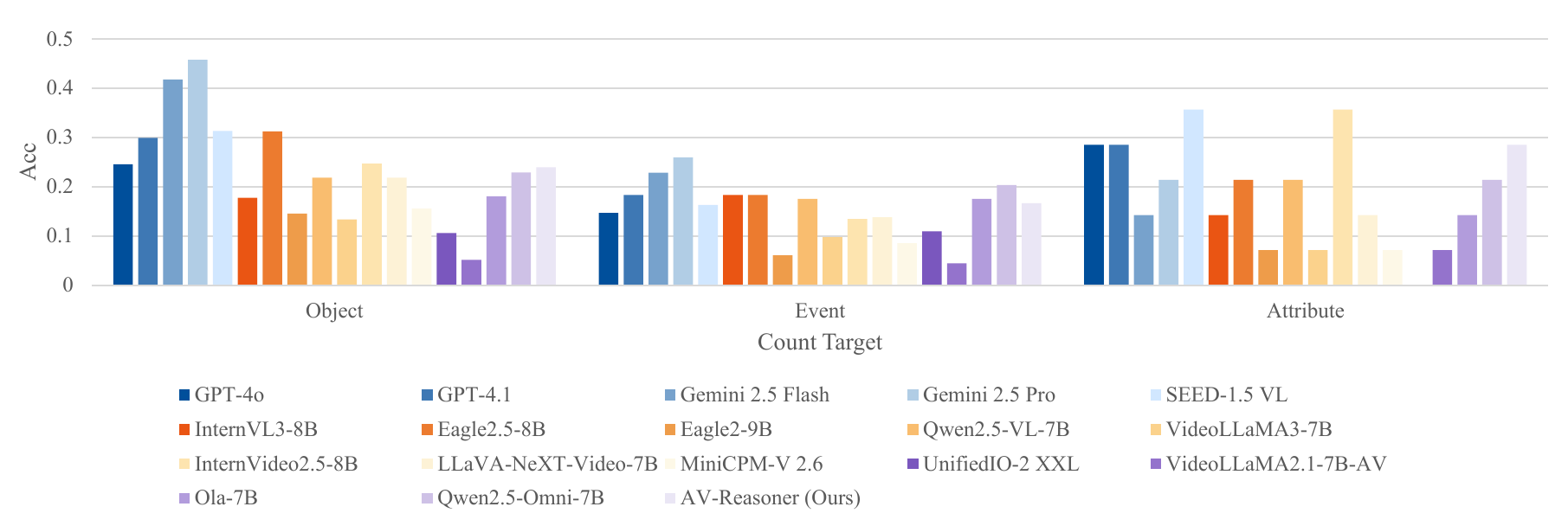}
    \caption{Model accuracy across different count targets.}
    \label{fig:count_target_acc}
\end{figure}

Fig.~\ref{fig:count_target_acc} highlights a clear trend across both close-source and open-source MLLMs: models consistently perform better on object counting tasks compared to event and attribute counting. Most models demonstrate a noticeable advantage when counting concrete, visually grounded entities like objects. This performance gap is especially prominent in open-source models, where accuracy on event and attribute targets often drops substantially.

The relative ease of object counting can be attributed to the more direct visual correspondence between input and target. Objects are typically well-defined spatially, consistently annotated in vision-language pretraining data, and often associated with discrete visual regions. In contrast, events may unfold over time and require temporal reasoning, while attributes tend to be abstract, context-dependent, or even implicit, making them harder to detect and quantify reliably.

These findings suggest that current MLLMs are more adept at processing perceptually salient elements. Addressing the challenges posed by more abstract or temporally extended targets like events and attributes may require stronger temporal modeling, better multimodal alignment, or targeted supervision in future model designs.

\subsubsection{Model Performance across Different Counting Numbers}
\begin{figure}[t]
    \centering
    \includegraphics[width=\linewidth]{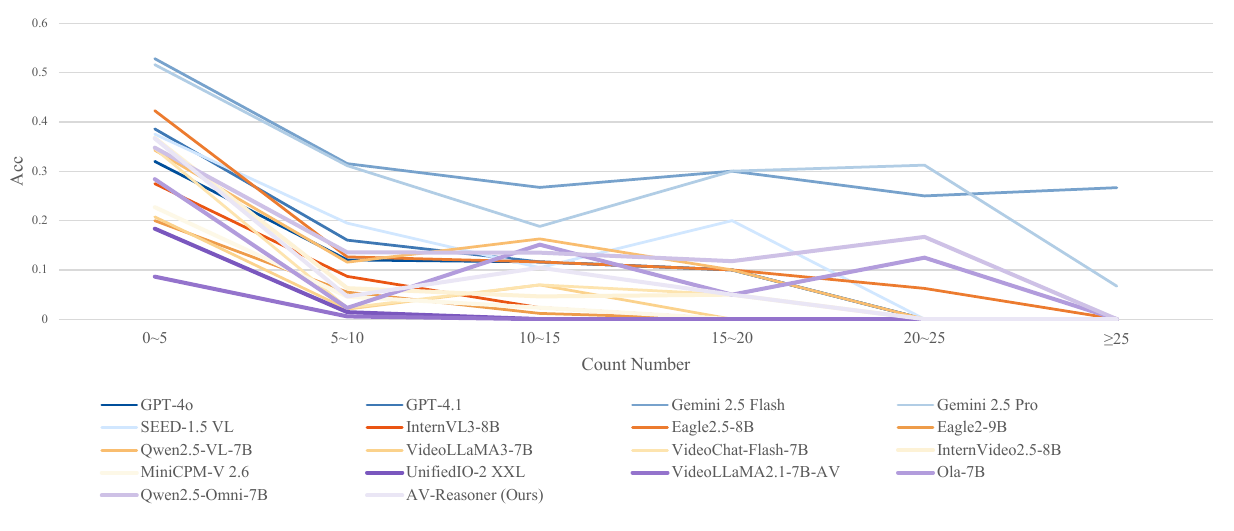}
    \caption{Model accuracy across different count ranges.}
    \label{fig:count_range_acc}
\end{figure}

As shown in the Fig.~\ref{fig:count_range_acc}, model performance clearly declines as the count range increases. Open-source models exhibit relatively high accuracy when the count is $\leq$5, but their performance becomes more erratic and less reliable when the count exceeds 5.
\subsubsection{Model Performance across Different Query Modalities}
\begin{figure}[t]
    \centering
    \includegraphics[width=\linewidth]{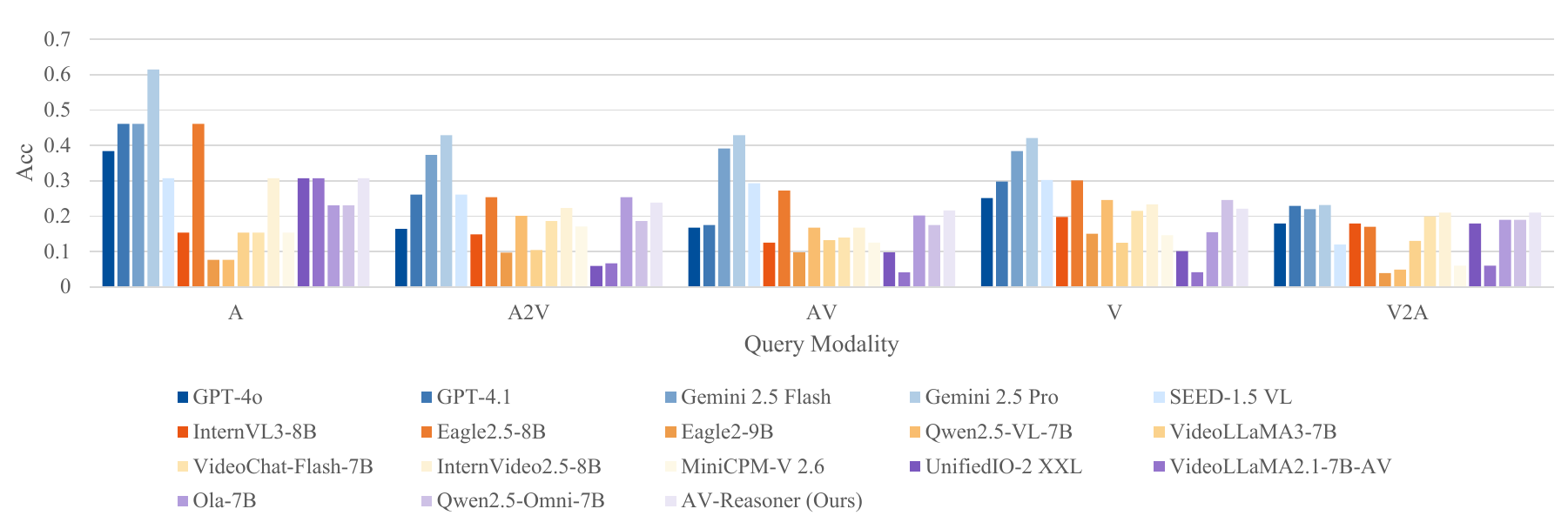}
    \caption{Model accuracy across different query modalities.}
    \label{fig:q_acc}
\end{figure}

As shown in Fig.~\ref{fig:q_acc}, model performance varies significantly across different query modalities. When the query modality is audio (A), most models—especially omni-MLLMs like Qwen2.5-Omni-7B and Ola-7B achieve relatively high accuracy and low error, indicating strong alignment with audio inputs. However, performance drops noticeably in cross-modal settings such as A2V and V2A, where both MAE and RMSE increase substantially. Notably, V2A and AV emerges as the most challenging configuration across all metrics. Overall, omni-MLLMs (purple) show clear advantages over VLMs (yellow) in audio-involved queries, but this advantage diminishes in purely visual or AV scenarios.

\subsubsection{Model's White-box Evaluation Performance across Different Counting Targets}
\begin{figure}[t]
    \centering
    \includegraphics[width=\linewidth]{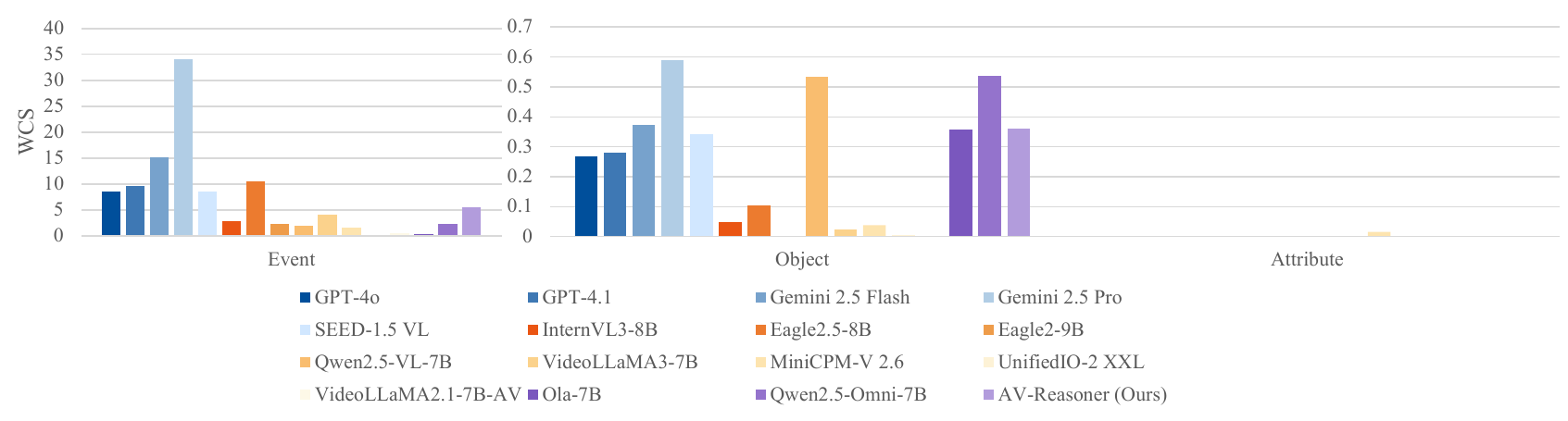}
    \caption{Model's White-box Evaluation Performance across Different Counting Targets.}
    \label{fig:accW}
\end{figure}
As shown in Fig.~\ref{fig:accW}, in the white-box evaluation, most models demonstrate relatively strong reasoning counting abilities in the event counting, but perform poorly in the object and attribute counting. Notably, only MiniCPM-V 2.6~\cite{yao2024minicpm-v} achieves a score in the attribute counting task. This suggests that current models are relatively proficient in temporal grounding, where understanding the sequence and timing of events is crucial, but perform poorly in spatial grounding, which requires precise localization and differentiation of objects or attributes.

\section{Experiment Details}
\subsection{Experimental Setups}
All experiments are conducted using the TRL framework on 8 NVIDIA A100 GPUs (each with 80GB memory). The base model used is Ola-Omni-7B. Training was performed with bf16 enabled to improve memory efficiency and computational performance.

We adopted the GRPO algorithm for fine-tuning. During the rollout phase, the generation temperature was set to 1.0, the maximum number of generated tokens is 256 or 1024 according to different tasks, and each sample was expanded into 8 rollout trajectories. The KL divergence penalty coefficient $\beta$ was set to 0.1 to ensure controlled deviation from the initial policy.
\subsection{Detailed Training Settings}
Tab.~\ref{tab:training_strategy} summarizes the training configurations across different stages. The curriculum-based RL process is divided into three subtasks: QA, grounding, and counting, each trained with its corresponding dataset. To mitigate forgetting and maintain performance across tasks, we adopt a stage review mechanism (SRM), which involves mixing a portion of previously seen samples during training. The number of such samples is shown in parentheses.
\begin{table*}[t]
    \centering
    \caption{Our detailed training settings. Data indicated in parentheses refers to that used in the SRM.}

    \setlength{\tabcolsep}{10pt}
    \renewcommand{\arraystretch}{1.2}
    \resizebox{1.0\linewidth}{!}{
    \begin{tabular}{@{}ll|ccccc@{}}
    \toprule
    & & \textbf{Cold-start SFT} & \textbf{Curriculum-based RL (QA)} & \textbf{Curriculum-based RL (Grounding)} &\textbf{Curriculum-based RL (Counting)}&\textbf{Full-task RL}\\ 
    \midrule 
    \multirow{2}{*}{Data}
    & \textbf{Dataset} & AVTG+ARIG+Counting &  AVQA  &  AVTG+ARIG (AVQA) & Counting (AVTG+ARIG+AVQA) & AVQA+AVTG+ARIG+Counting 
    \\
    & \#Samples & 78K & 72K & 72K (14K) & 6K (1K) & 10K \\
    \midrule
    \multirow{2}{*}{Training}
    & \textbf{Thinking} & \ding{55}& \multicolumn{4}{|c}{\ding{51}}\\ 
     & \textbf{Max New Tokens} & - & 256 & \multicolumn{3}{|c}{1024}\\ 
    \bottomrule
    \end{tabular}}
    \label{tab:training_strategy}
    \end{table*}

\section{Evaluation on More Benchmarks}
The performance of our model is also evaluated on the AVQA and AVE benchmarks. The test results are shown in the Tab.~\ref{tab:MORE_BENCH}.
\begin{table*}[t]
\centering
\caption{Performance comparison across AVQA and AVE Benchmarks.}
\label{tab:MORE_BENCH}
\setlength\tabcolsep{8pt}
\renewcommand{\arraystretch}{1.2}
\resizebox{\textwidth}{!}{
\begin{tabular}{l|cccccc}
\toprule
Benchmark &  MM-Pyramid~\cite{yu2022mm}&MEERKAT~\cite{chowdhury2024meerkat} & PAVE~\cite{liu2025pavepatchingadaptingvideo} & Crab~\cite{du2025crab} & AV-Reasoner (Ours) & AV-Reasoner-Thinking (Ours) \\
\midrule
AVQA Acc (\%) &-& 87.17 & 93.80 & - & 93.02 & 93.17 \\
AVE Acc (\%)  &77.80& - & - & 80.15 & 82.86 & 81.26 \\
\bottomrule
\end{tabular}}
\end{table*}

\end{document}